\documentclass[runningheads]{llncs}
\usepackage{graphicx}

\usepackage{tikz}
\usepackage{comment} 
\usepackage{amsmath,amssymb} 
\usepackage{color}
\usepackage{url}
\usepackage{bm}

\begin{document}
\pagestyle{headings}
\mainmatter
\def\ECCVSubNumber{5346}  

\title{Neural Wireframe Renderer: Learning Wireframe to Image Translations} 

\titlerunning{Neural Wireframe Renderer}
%
\author{Yuan Xue \and
Zihan Zhou \and
Xiaolei Huang}
\authorrunning{Y. Xue et al.}
%
\institute{College of Information Sciences and Technology, \\ The Pennsylvania State University, University Park, PA 16802, USA}
\maketitle

\begin{abstract}
In architecture and computer-aided design, wireframes (\textit{i.e.}, line-based models) are widely used as basic 3D models for design evaluation and fast design iterations. However, unlike a full design file, a wireframe model lacks critical information, such as detailed shape, texture, and materials, needed by a conventional renderer to produce 2D renderings of the objects or scenes. In this paper, we bridge the information gap by generating photo-realistic rendering of indoor scenes from wireframe models in an image translation framework. While existing image synthesis methods can generate visually pleasing images for common objects such as faces and birds, these methods do not explicitly model and preserve essential structural constraints in a wireframe model, such as junctions, parallel lines, and planar surfaces. To this end, we propose a novel model based on a structure-appearance joint representation learned from both images and wireframes. In our model, structural constraints are explicitly enforced by learning a joint representation in a shared encoder network that must support the generation of both images and wireframes. Experiments on a wireframe-scene dataset show that our wireframe-to-image translation model significantly outperforms the state-of-the-art methods in both visual quality and structural integrity of generated images.

\end{abstract}

\section{Introduction}

\begin{figure}[t]
\centering
\includegraphics[width=0.95\linewidth]{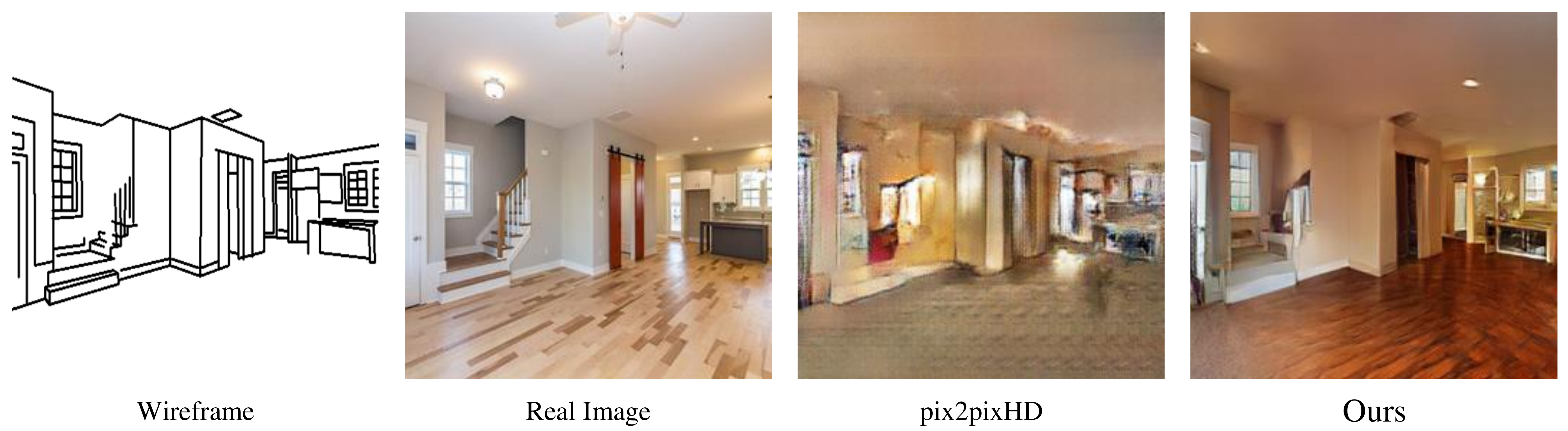}
\caption{Wireframe-to-image translation: given the wireframe of a man-made environment, the goal is to generate a photo-realistic rendering of the scene.}
\label{fig:teaser}
\end{figure}

Recently, driven by the success of generative adversarial networks (GANs)~\cite{goodfellow2014generative} and image translation techniques~\cite{isola2017image,zhu2017unpaired}, there has been a growing interest in developing data-driven methods for a variety of image synthesis applications, such as image style transfer~\cite{johnson2016perceptual,karras2018style}, super-resolution~\cite{ledig2017photo}, enhancement~\cite{zhang2017image}, text-to-image generation~\cite{zhang2017stackgan++}, domain adaption~\cite{HoffmanTPZISED18,MurezKKRK18}, just to name a few. In this work, we study a new image synthesis task, dubbed \emph{wireframe-to-image translation}, in which the goal is to convert a wireframe (\textit{i.e.}, a line-based skeletal representation)
of a man-made environment to a photo-realistic rendering of the scene (Fig.~\ref{fig:teaser}). In the fields of visual arts, architecture, and computer-aided design, the wireframe representation is an important intermediate step for producing novel designs of man-made environments. For example, commercial computer-aided design software such as AutoCAD allows designers to create 3D wireframe models as basic 3D designs for evaluation and fast design iterations.

In practice, for designers to quickly validate their design and obtain feedback from customers, it is often necessary to convert such a 3D wireframe into a photo-realistic rendering of the scene in real-time. However, compared to a full design file, a wireframe model lacks information needed by a conventional rendering engine, such as detailed shape and texture information as well as materials. 

\begin{figure}[t]
\centering
\includegraphics[width=0.90\linewidth]{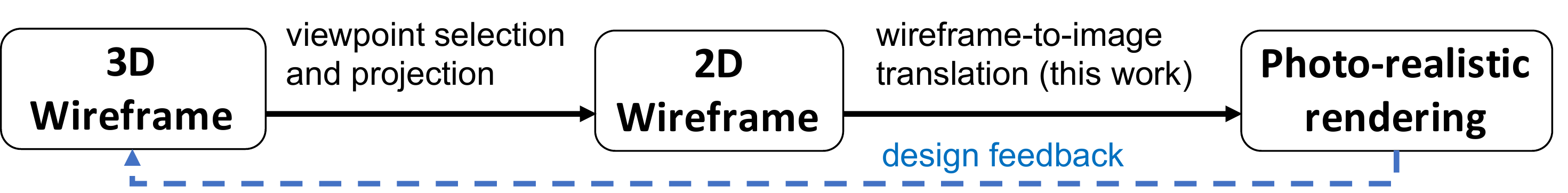}
\caption{Our envisioned workflow for design applications.}
\label{fig:workflow}
\end{figure}

In this paper, we address the need for generating 2D renderings from wireframe models in design applications. Fig.~\ref{fig:workflow} illustrates our envisioned workflow, in which a 3D wireframe is first projected to the image plane given a viewpoint chosen by the user. Then, a deep network is trained to convert the 2D wireframe into a realistic scene image.
Note that, compared to edge maps and sketches, wireframes contain precise information that encodes 3D geometric structure such as salient straight lines and junctions while being more sparse and ignoring lines due to planar texture. As such, an image generated given a wireframe input should respect the geometric constraints encoded in it, and should have pixel-level correspondence around straight lines and junctions where lines intersect. This requirement arises from the fact that human perception of 3D is highly dependent on recognizing structures like those encoded in a wireframe; even small violations of such constraints would make a generated image look unnatural. 

\begin{figure}[t]
\begin{tabular}{ccc}
\hspace{-1mm}\includegraphics[height = 0.70in]{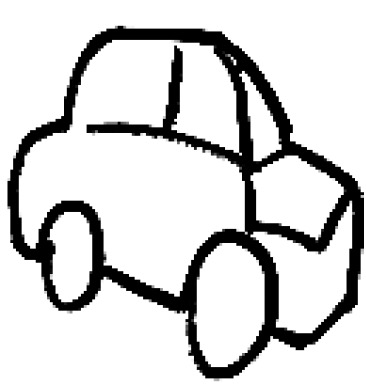}
\includegraphics[height = 0.70in]{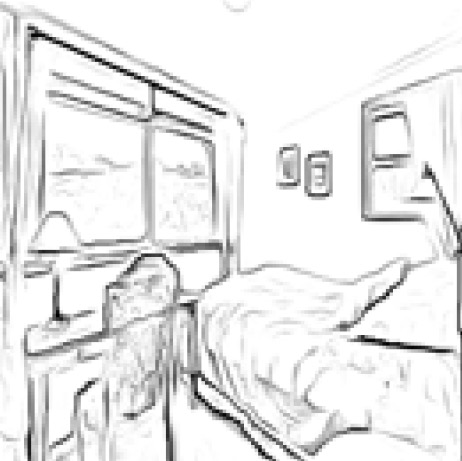} &
\hspace{-1mm}\includegraphics[height = 0.70in]{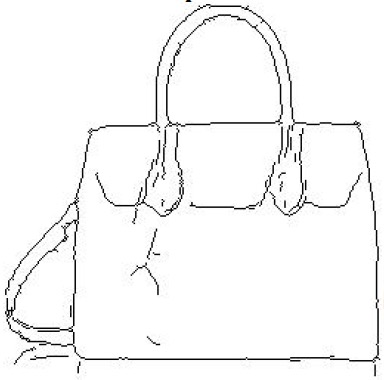}
\includegraphics[height = 0.70in]{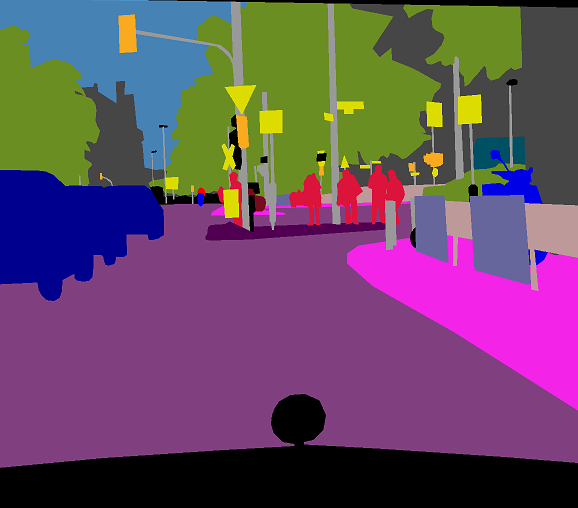} &
\hspace{-1mm}\includegraphics[height = 0.70in]{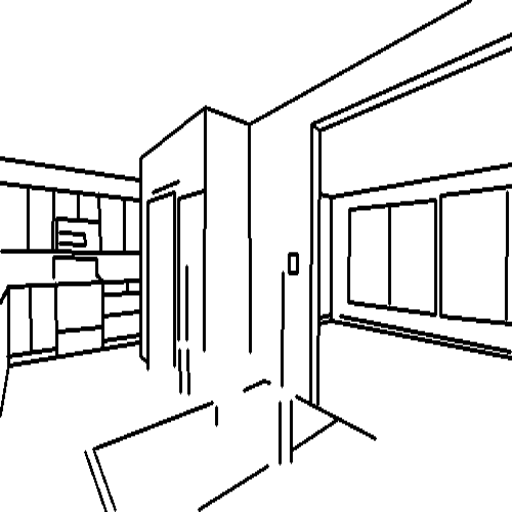}
\includegraphics[height = 0.70in]{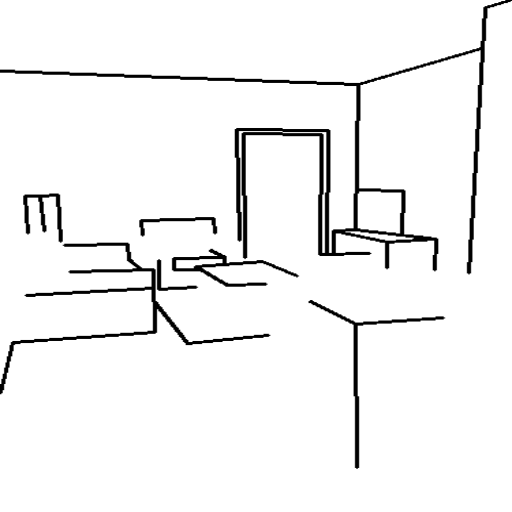} \\
(a) Sketches~\cite{SangkloyLFYH17,LuWTT18} &
(b) Edge and semantic maps~\cite{isola2017image} &
(c) Wireframes
\end{tabular}
\caption{Comparison of different input modalities in image translation tasks. Compared with other modalities, wireframes contain more prominent structural constraints in terms of straight lines and their relationships (\textit{\textit{e.g.}}, junctions, parallelism, and orthogonality), while being sparser and containing less semantic information.}
\label{fig:modality}
\end{figure}

State-of-the-art image translation models such as pix2pixHD~\cite{wang2018high} have difficulty in generating images that preserve structures such as straight lines and their intersections (Fig.~\ref{fig:teaser}).  This may be due to that these models are designed for other types of input modalities, as illustrated in Fig.~\ref{fig:modality}.  Inputs that are semantic segmentation maps emphasize object instance- or part-level correspondence rather than pixel-level correspondence; scribbles in free-hand sketches usually do not strictly map to lines or curves in photographic images; and edges often do not contain complete and accurate structure information and make no distinction between salient structural lines and planar texture-induced lines.

In this work, we propose a structure-appearance joint representation learning scheme that utilizes a paired wireframe-image dataset to learn to generate images with \emph{structural integrity}. Our assumption is that there exists a shared latent space for encoding both structural and appearance constraints of a scene. Accordingly, we design our wireframe-to-image translation model to include one encoder and two decoders (see Fig.~\ref{fig:framework_translation}). The encoder encodes the input wireframe to a joint representation, the wireframe decoder reconstructs the original wireframe from the joint representation, and the scene decoder transforms the representation into a photo-realistic indoor image.  Further, the jointly generated wireframe-image pairs are used to train a cGAN-like~\cite{mirza2014conditional} discriminator, which takes the generated pairs as fake samples and groundtruth wireframe-image pairs as real samples. Such a design enables us to better preserve structural integrity and pixel-level correspondences in two ways. \emph{First}, the encoder together with the wireframe decoder branch can be regarded as an autoencoder for wireframes, which helps  enforce precise pixel-level correspondences for salient structures.
\emph{Second}, the cGAN-like discriminator provides an adversarial loss that can help train the model to adaptively learn the difference between the reconstructed wireframe-image pairs and groundtruth pairs.

We demonstrate the effectiveness of our proposed model by conducting extensive experiments on a dataset of various indoor scene images with ground truth wireframe annotations~\cite{huang2018learning}. As shown in Fig.~\ref{fig:teaser} and results in Section~\ref{sec:experiments},
by introducing a joint representation learning scheme combined with both adversarial loss and perceptual loss, our proposed wireframe renderer generates images that not only have higher visual realism than prior arts~\cite{chen2017photographic,isola2017image,wang2018high,park2019semantic}, but also adhere much better to structural constraints encoded in the input wireframes.

To summarize, the main contributions of our work are:
\begin{itemize}
\item We propose a supervised image to image translation model which generates realistic image renderings from wireframe inputs. The architecture including a novel structure-appearance joint representation and multiple loss functions for the end-to-end network are carefully designed to ensure that the generated synthetic images adhere to wireframe structural constraints.
\item To the best of our knowledge, we are the first to conduct wireframe-to-image translation experiments for high-fidelity indoor scene rendering using a challenging indoor scene wireframe dataset.
Both quantitative and qualitative results of our experiments indicate the superiority of our proposed method compared with previous state-of-the-art methods.
\end{itemize}

\section{Related Work}

\smallskip
\noindent{\bf Wireframe Parsing.} Several methods have been developed recently to extract wireframes from images~\cite{huang2018learning,xue2019learning,zhou2019end,Zhou19}. In this paper, we study the inverse problem of translating wireframes to photo-realistic images. 

\smallskip
\noindent{\bf Generative Adversarial Networks.}
Generative adversarial networks (GANs) \cite{goodfellow2014generative}, especially the conditional GANs~\cite{mirza2014conditional}, have been widely used in image synthesis applications such as text-to-image generation~\cite{zhang2017stackgan++} and image-to-image translation~\cite{isola2017image,zhu2017unpaired}. 
However, training GANs is known to be difficult and often requires a large training set in order to generate satisfactory results. Some attempts have be made to stabilize the GAN training~\cite{gulrajani2017improved,mao2017least}, as well as use coarse-to-fine generation to get better results~\cite{karras2017progressive,zhang2017stackgan++}. 
One work that explores structure information in GAN training is~\cite{wang2016generative}. It utilizes RGB-D data and factorizes the image generation process into synthesis of a surface normal map and then the conditional generation of a corresponding image. 

\smallskip
\noindent{\bf Supervised Image-to-Image Translation.}
The line of research that most closely relate to our work is supervised image-to-image translation, in which input-output image
pairs are available during training. Prior work~\cite{chen2017photographic,isola2017image,wang2018high} has been focusing on leveraging different losses to generate high-quality output images. While pixel-wise losses, such as the $\ell_1$ loss, are the most natural choices, using $\ell_1$ loss alone has been shown to generate blurry images~\cite{isola2017image,johnson2016perceptual}. To mitigate the problem, Isola \textit{et~al.}~\cite{isola2017image} uses a combination of $\ell_1$ loss and a conditional adversarial loss. To avoid the instability of adversarial training, Chen and Koltun~\cite{chen2017photographic} implement a cascaded refinement network trained via feature matching based on a pre-trained visual perception network. Recently, the perceptual loss~\cite{dosovitskiy2016generating} has been shown to be effective in measuring the perceptual similarity between images~\cite{zhang2018unreasonable}.
Wang~\textit{et~al.}~\cite{wang2018perceptual} integrates the perceptual adversarial loss and the generative adversarial loss to adaptively learn the discrepancy between the output and ground-truth images. Combining the merits from previous works, Wang~\textit{et~al.}~\cite{wang2018high} generate high quality images with coarse-to-fine generation, multi-scale discriminators, and an improved adversarial loss. 

Other works focus on improving the performance for a certain input modality. For \emph{semantic maps}, Qi~\textit{et~al.}~\cite{qi2018semi} first retrieve segments from external memory, then combine the segments to synthesize a realistic image.
Liu~\textit{et~al.}~\cite{liu2019learning} predict convolutional kernels from semantic labels and use a feature-pyramid semantics-embedding discriminator for better semantic alignment. Park~\textit{et~al.}~\cite{park2019semantic} modulate the normalization layer with learned parameters to avoid washing out the semantic information. For \emph{sketches}, Sangkloy~\textit{et~al.}~\cite{SangkloyLFYH17} generate realistic images by augmenting the training data with multiple sketch styles; SketchyGAN~\cite{chen2018sketchygan} improves the information flow during training by injecting the input sketch at multiple scales; Lu~\textit{et~al.}~\cite{LuWTT18} use sketch in a joint image completion framework to handle the misalignment between sketches and photographic objects.

\smallskip
\noindent{\bf Joint Representation Learning.}
For applications that involve two or more variables, the traditional one-way mapping of GANs may be insufficient to guarantee the correspondence between the variables. Conditional GANs~\cite{mirza2014conditional,chen2016infogan,odena2017conditional,choi2018stargan} learn to infer one variable from another in both directions. ALI~\cite{dumoulin2016adversarially}, CycleGAN~\cite{zhu2017unpaired}, and their variants (\textit{\textit{e.g.}},~\cite{li2017alice,kim2017learning,yi2017dualgan}) learn the cross-domain joint distribution matching via bidirectional mapping of two examples. 

In unsupervised image-to-image translation, several works~\cite{huang2018multimodal,liu2017unsupervised,LeeTHSY18} propose to map images from multiple domains to a joint latent space. To further learn instance level correspondences, DA-GAN~\cite{ma2018gan} incorporate a consistency loss in the latent space between the input and output images. However, due to the lack of paired training data, it is hard for these methods to generate outputs that match all the details (\textit{e.g.}, semantic parts) in the input images.

When paired data is available, learning a joint representation has been proved to be an effective way to capture the correspondences. To promote instance awareness in unsupervised image translation, InstaGAN~\cite{MoCS19} simultaneously translates image and the corresponding segmentation mask. Recent work on domain adaption~\cite{ChenLCG19} jointly predict segmentation and depth maps in order to better align the predictions of the task network for two domains.

\section{Methodology}
\begin{figure}[t]
\centering
\includegraphics[width=0.99\linewidth]{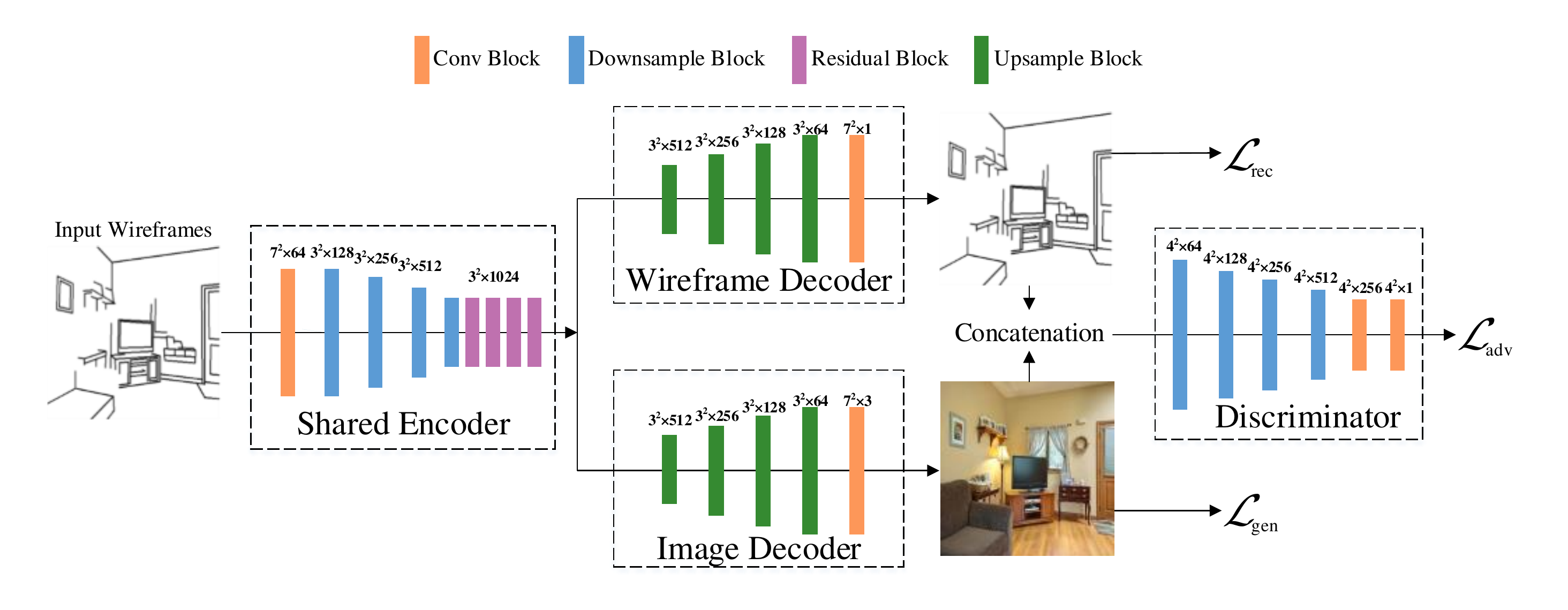}
   \caption{Network architecture of our wireframe-to-image translation model. The numbers above each block indicate the kernel size and the output channel number.}
\label{fig:framework_translation}
\end{figure}

In this work, we propose to add an intermediate step in the image synthesis process to improve structural integrity and pixel-level correspondence. Specifically, we learn a structure-appearance joint representation from the input wireframe, and use the joint representation to simultaneous generate corresponding scene images and reconstructed wireframes as output. As shown in Fig.~\ref{fig:framework_translation}, The overall pipeline of our wireframe renderer consists of an encoder, a wireframe decoder, a scene image decoder, and a discriminator.

In the following, we introduce the theoretical background and architecture of our model in Section~\ref{sec:method:embedding}, and discuss implementation details in Section~\ref{implementation_translation}.

\subsection{Learning Joint Representation for Wireframe-to-Image Translation}
\label{sec:method:embedding}
Formally, we measure the uncertainty of generating the correct wireframe from a joint representation of wireframe and scene image using \emph{Conditional Entropy}. The conditional entropy of an input wireframe $x$ conditioned on its corresponding joint representation $e$ is defined as 
\begin{equation}
H(x|e) = \mathbb{E}_{x \sim P(x|e)} \log  P(x|e),
\end{equation}
where $e \sim \hat{Q}(x,y) $ follows an estimated joint distribution $\hat{Q}$ of wireframe $x$ and indoor scene image $y$, and is computed by an encoder network $\text{Enc}$. Under a supervised training scenario with paired wireframe and scene image, for simplicity, we assume that the mapping from $x$ to $e$ is deterministic so that $e = \text{Enc}(x)$ is a joint representation of $x$ and $y$. Since the mapping from $e$ to $x$ should also be deterministic when $e$ contains a certain input, we have $H(x|e= \text{Enc}(x)) = 0$.

Since we do not have the ground truth distribution of $P(x|e)$, we approximate it with a decoder network $\text{Dec}_w$ for reconstructing the wireframe from the joint representation. The conditional entropy of $\text{Dec}_w$ is
\begin{equation}
\begin{split}
\mathcal{L}_{ce}&=\mathbb{E}_{\hat{x} \sim P(x|e)} \log  \text{Dec}_w(\hat{x}|e) \\&= H(x|e) + \text{KL}(P(x|e)||\text{Dec}_w(\hat{x}|e)) \geq H(x|e) = 0.
\end{split}
\end{equation}
Thus, minimizing the conditional entropy is equivalent to reducing the KL divergence between the decoder and the ground truth posterior. To approximate the $\mathcal{L}_{ce}$, given a mini-batch of $N$ wireframes $x_n$, we define the wireframe reconstruction objective as
\begin{small}
\begin{equation}
\begin{split}
\min\limits_{\theta, \theta_w}\mathcal{L}_{\text{rec}}= & \frac{1}{N}\sum_{n=1}^N \Big(\alpha_w || x_n - \text{Dec}_w(\text{Enc}(x_n)) ||_{1} + \beta_w \text{MS-SSIM}(x_n,\text{Dec}_w(\text{Enc}(x_n)) \Big),
\end{split}
\end{equation}
\end{small}
where $\theta, \theta_w$ are the parameters of encoder and wireframe decoder, respectively. The first term is the $\ell_1$ distance between the original wireframe and the reconstructed wireframe. The second term is the Multiscale Structural Similarity (MS-SSIM) loss to compensate the $\ell_1$ distance, as MS-SSIM is more perceptually preferable. More details of MS-SSIM can be found in~\cite{wang2003multiscale}. $\alpha_w$ and $\beta_w$ are scaling factors to balance the two loss terms.

In addition to the decoder branch that reconstructs the wireframe, we have another decoder $\text{Dec}_s$ that generates the corresponding scene image from the learned joint representation.  By having the two decoder branches share the same encoder, the encoder network is forced to learn both structure and appearance information as well as their correspondence so that the generated image can have better structural alignment with the reconstructed wireframe. 

Given a mini-batch of $N$ wireframes $x_n$ and corresponding scene images $y_n$, we define the objective for scene generation as
\begin{small}
\begin{equation}
\min\limits_{\theta, \theta_s}\mathcal{L}_{\text{gen}} = \frac{1}{N}\sum_{n=1}^N \Big(\alpha_s || y_n - \text{Dec}_s(\text{Enc}(x_n)) ||_{1} + \beta_s  D_{\text{perc}}(y_n,\text{Dec}_s(\text{Enc}(x_n))\Big),
\end{equation}
\end{small}
where the scene decoder network is parameterized by $\theta_s$. The perceptual loss $D_{\text{perc}}$ is defined as 
\begin{equation}
D_{\text{perc}}(y,\hat{y}) = \sum_{l} \frac{1}{H_lW_l} ||\phi_l(y) - \phi_l(\hat{y})||_{2}^{2},
\label{Eq:perception}
\end{equation}
where $\phi_l$ is the activations of the $l$th layer of a perceptual network with shape $C_l \times H_l \times W_l$. In our experiments, we use the 5 convolutional layers from VGG16~\cite{simonyan2014very} pre-trained on ImageNet~\cite{russakovsky2015imagenet} to extract visual features, and unit-normalize the activations in the channel dimension as in~\cite{zhang2018unreasonable}.

Further, we propose an adversarial loss~\cite{goodfellow2014generative} to adaptively learn the difference between the reconstructed wireframe/generated image and the groundtruth. Denote $\hat{x}$ and $\hat{y}$ as the reconstructed wireframe and generated scene image, the adversarial objective is
\begin{equation}
\max\limits_{\theta_d}\min\limits_{\theta, \theta_w, \theta_s}\mathcal{L}_{\text{adv}} =  \mathbb{E}_{x,y}\log\sigma(\text{Dis}({x},{y})) + \mathbb{E}_{x,y}\log( 1 - \sigma(\text{Dis}(\hat{x},\hat{y}))),
\end{equation}
where $\sigma(\cdot)$ is the sigmoid function and $\theta_d$ represents the parameters of the conditional discriminator network, $\text{Dis}$. For simplicity, we omit the representations such as $x \sim P_x$ in all adversarial objectives. 

Therefore, the full objective for end-to-end training of our model is
\begin{equation}
\max\limits_{\theta_d} \min\limits_{\theta, \theta_w, \theta_s}\mathcal{L} =  \mathcal{L}_{\text{rec}} + \mathcal{L}_{\text{gen}} + \lambda \mathcal{L}_{\text{adv}},
\end{equation}
where $\lambda$ is another scaling factor to control the impact of the adversarial loss.

\subsection{Implementation Details}
\label{implementation_translation}
In our wireframe renderer model\footnote{Code available at \url{https://github.com/YuanXue1993/WireframeRenderer}}, the encoder network consists of $5$ convolution blocks. The first block uses $7\times7$ convolution kernels with stride $1$ and reflection padding $3$. The remaining $4$ downsample blocks have kernel size $3$, stride $2$ and reflection padding $1$. Each convolutional layer is followed by one batch normalization~\cite{ioffe2015batch} layer and one LeakyReLU~\cite{maas2013rectifier} activation. The last downsample block is followed by $4$ residual blocks~\cite{he2016deep} with $3\times 3$ convolution and ReLU activation.

The decoder consists of $4$ upsample blocks. To avoid the characteristic artifacts introduced by the transpose convolution~\cite{odena2016deconvolution}, each upsample block contains one $3\times3$ sub-pixel convolution~\cite{shi2016real} followed by batch normalization and ReLU activation. The last block uses a $7\times7$ convolution followed by a tanh activation without normalization. The two decoder networks have similar architecture except in the last layer where the outputs have different channel dimensions.

We follow~\cite{isola2017image} and use the PatchGAN~\cite{long2015fully} discriminator for adversarial training. We use LSGAN~\cite{mao2017least} for stabilizing the adversarial training. The scaling factors in our final model are $\alpha_w = 1, \beta_w = 1, \alpha_s = 15, \beta_s = 4$ and $\lambda = 1$. These values are determined through multiple runs of experiments. The training is done using Adam optimizer~\cite{kingma2014adam} with initial learning rate $2e-3$. The learning rate is decayed every $30$ epochs with rate $0.5$. The batch size is $16$ and the maximum number of training epochs is $500$.

All training images are first resized to $307 \times 307$, then randomly cropped to $256 \times 256$. A random horizontal flipping and random adjustment of brightness, contrast and saturation are applied for data augmentation. During inference, all images are re-scaled to $256 \times 256$ with no further processing.

\section{Experiments} \label{sec:experiments}
\subsection{Experiment Settings}

\noindent{\bf Dataset.} The wireframe dataset~\cite{huang2018learning} consists of 5,462 images of man-made environments, including both indoor and outdoor scenes, and manually annotated wireframes. Each wireframe is represented by a set of junctions, a set of line segments, and the relationships among them. Note that, unlike general line segments, the wireframe annotations consider structural elements of the scene only. Specifically, line segments associated with the scene structure are included, whereas line segments associated with texture (\textit{\textit{e.g.}}, carpet), irregular or curved objects (\textit{\textit{e.g.}}, humans and sofa), and shadows are ignored. Thus, to translate the wireframe into a realistic image, it is critical for a method to handle incomplete information about scene semantics and objects.

As we focus on the indoor scene image generation task in this paper, we filter out all outdoor or irrelevant images in the dataset. This results in 4,511 training images and 422 test images. The dataset contains various indoor scenes such as bedroom, living room, and kitchen. It also contains objects such as humans which are irrelevant to our task. The limited size and the scene diversity of the dataset make the task of generating interior design images even more challenging.

\medskip
\noindent{\bf Baselines.} We compare our image translation models with several state-of-the-art models, namely the Cascaded Refinement Network (CRN)~\cite{chen2017photographic}, pix2pix~\cite{isola2017image}, pix2pixHD~\cite{wang2018high}, and SPADE~\cite{park2019semantic}. For fair comparison, we adapt from the authors' original implementations wherever possible. For CRN, we use six refine modules, starting from $8\times8$ all the way up to $256\times256$. For pix2pix model, we use UNet~\cite{ronneberger2015u} backbone model as in the original paper. We decrease the weight of pixel loss from 100 to 50 since the original weight fails to generate any meaningful results. For pix2pixHD model, we use two discriminators with different scales and the discriminator feature matching loss combined with the GAN loss. Since there is no instance map available for our problem, we train the pix2pixHD model with wireframes only. For SPADE, we use at most 256 feature channels to fit in the single GPU training.

Besides, to verify the benefit of joint representation, we also train a variant of our method in which we remove the wireframe decoder branch. All the other components in the network are the same as our full model and we train it with the same image generation loss and adversarial loss for wireframe-to-image translation. For all baseline models involving adversarial training, since there is no wireframe predicted, the generated images are paired with their input wireframes as the input to the discriminator.

\subsection{Qualitative Comparisons}

\begin{figure}[t]
\centering
\includegraphics[width=0.99\linewidth]{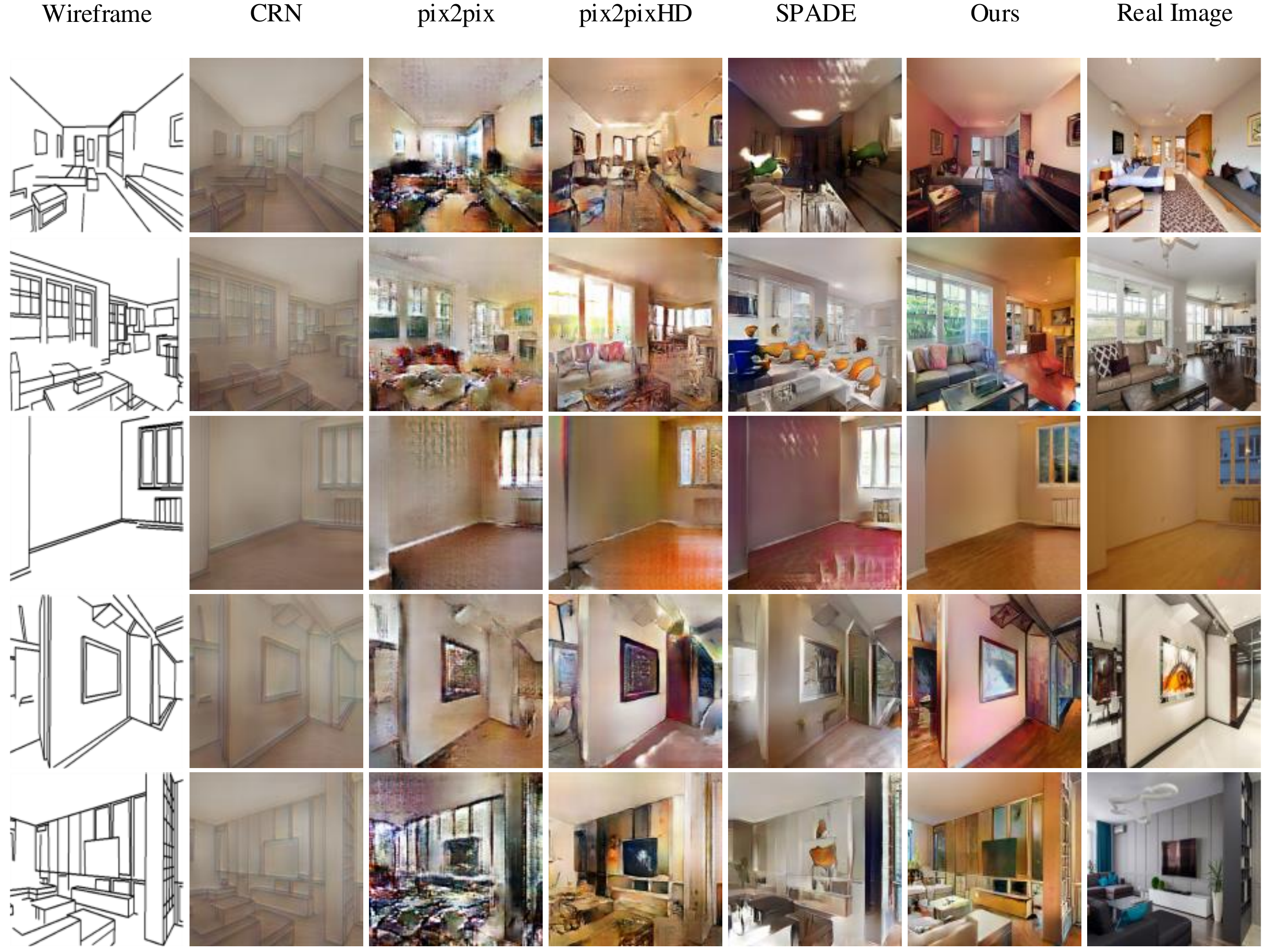}
\caption{Qualitative comparison for image translation models on the test set. Each row represents one wireframe/image pair and each column represents one model. The input wireframes and corresponding groundtruth images are included as references.}
\label{fig:result_translation}
\end{figure}

The qualitative comparisons for the translation models are shown in Fig.~\ref{fig:result_translation}. We first note that the CRN model trained on the wireframe dataset fails to generate meaningful results, despite that we have experimented with different hyper-parameter settings. 
One possible reason is that the CRN is originally designed for image synthesis based on semantic layouts. However, the wireframe itself contains little semantic information (\textit{\textit{e.g.}}, object categories), thus the model has to infer such information from the structure information presented in the wireframe. Moreover, CRN model is the only model which does not use adversarial training. This may suggest that adversarial training is important in the wireframe-to-image translation task. 

Except for the CRN, all other models are able to generate meaningful synthetic images. However, in the images generated by pix2pix and pix2pixHD, structural integrity is not always well preserved. In general, the generated images of these models cannot align well with the input wireframes, especially when structure information is complicated (\textit{e.g.}, the furniture areas in the first and second rows of Fig.~\ref{fig:result_translation}). Further, these methods generate noticeable artifacts in regions where structure information is sparse (\textit{e.g.}, the white walls in the third row of Fig.~\ref{fig:result_translation}). 
For SPADE~\cite{park2019semantic}, structural information is better preserved, but the results contain more artifacts than those of pix2pixHD and appear to be less realistic (\textit{e.g.}, artifacts in the first, second, and fifth rows of Fig.~\ref{fig:result_translation}).
In contrast, our model generates images with best quality among all models and preserves the structure and correspondence very well. Compared with the real images in the test set, the synthetic images of our final model are almost photo-realistic.

\subsection{Quantitative Evaluations}

\noindent{\bf FID, LPIPS, and SSIM scores.} We first report results based on various standard metrics for image synthesis. Fr\'echet inception distance (FID)~\cite{heusel2017gans} is a popular evaluation metric for image synthesis tasks, especially for GAN models.
FID features are extracted from an Inception-v3~\cite{szegedy2016rethinking} model pre-trained on ImageNet. Since the dataset contains various indoor scenes, we use the pre-trained model without fine-tuning. Lower FID score indicates a better generation result.

For our task, since we have the ground truth images associated with the input wireframes, we also calculate paired LPIPS and SSIM scores between the synthetic images and the real images. The learned perceptual image patch similarity (LPIPS)~\cite{zhang2018unreasonable} is essentially a perceptual loss. It has been shown to have better agreement with human perception than traditional perceptual metrics such as SSIM~\cite{wang2004image} and PSNR. We use Eq.~\eqref{Eq:perception} to calculate the perceptual distance between the synthetic image and the real image. Note that in our experiments we calculate the perceptual distance instead of the similarity, thus the lower the LPIPS score, the better quality of the generated images. The feature extractor is a pre-trained VGG16 model as in our model training.

In Table~\ref{Tb:translation}(left), we report results of all methods except for CRN, since CRN fails to generate meaningful results. As one can see, pix2pixHD outperforms pix2pix in all metrics.  Compared with the pix2pix, pix2pixHD adopts multi-scale discriminators and use the adversarial perceptual loss, leading to better performance in the image translation task. However, since the training dataset in our experiments has a limited size, a perceptual loss learned by adversarial training may not work as well as a perceptual loss computed by a pre-trained feature extractor. As shown in Table~\ref{Tb:translation}, our model without the joint representation learning achieves better performance than the pix2pixHD model. 

\begin{table}[t]
\centering
\caption{Quantitative evaluation on the wireframe-to-image translation task. {\bf Left:} Standard image synthesis metrics. For SSIM, the higher the better; For FID and LPIPS, the lower the better. {\bf Right:} Wireframe parser scores using~\cite{zhou2019end}. For sAP scores, the higher the better.}
\label{Tb:translation}
\begin{tabular}{l|c|c|c}
Method &  FID$\downarrow$ & LPIPS$\downarrow$ & SSIM$\uparrow$  \\
\hline\hline
pix2pix~\cite{isola2017image}  & $186.91$ & $3.34$ & $0.091$ \\
\hline
pix2pixHD~\cite{wang2018high}  & $153.36$ & $3.25$ & $0.080$ \\
\hline
SPADE~\cite{park2019semantic}  & $93.90$ & $2.95$ & $0.086$ \\
\hline
Ours w/o JR  & $97.49$ & $2.85$ & $0.092$ \\
\hline
Ours  & $\bm{70.73}$ & $\bm{2.77}$ & $\bm{0.102}$ \\
\hline
\end{tabular}
\;\;
\begin{tabular}{l|c|c|c}
Method &  sAP$^5\uparrow$ & sAP$^{10}\uparrow$ & sAP$^{15}\uparrow$  \\
\hline\hline
pix2pix~\cite{isola2017image}  & $7.8$ & $10.0$ & $11.1$ \\
\hline
pix2pixHD~\cite{wang2018high}  & $10.6$ & $13.6$ & $15.1$ \\
\hline
SPADE~\cite{park2019semantic}  & $54.7$ & $58.1$ & $59.5$ \\
\hline
Ours w/o JR  & $26.7$ & $34.4$ & $37.5$ \\
\hline
Ours  & $\bm{60.1}$ & $\bm{64.1}$ & $\bm{65.7}$ \\
\hline
\hline
Real images & $58.9$ & $62.9$ & $64.7$ \\
\hline
\end{tabular}
\end{table}

Finally, our full model with the joint representation learning achieves the best performance across all metrics, as the images generated by the model better preserve the structure information encoded in the wireframes.

\medskip
\noindent{\bf Wireframe detection score.} Since the focus of this work is to preserve structure information in the wireframe-to-image translation task, an important and more meaningful evaluation metric would be whether we can infer correct wireframes from the generated images or not.

\begin{figure}[t]
\centering
\includegraphics[width=0.95\linewidth]{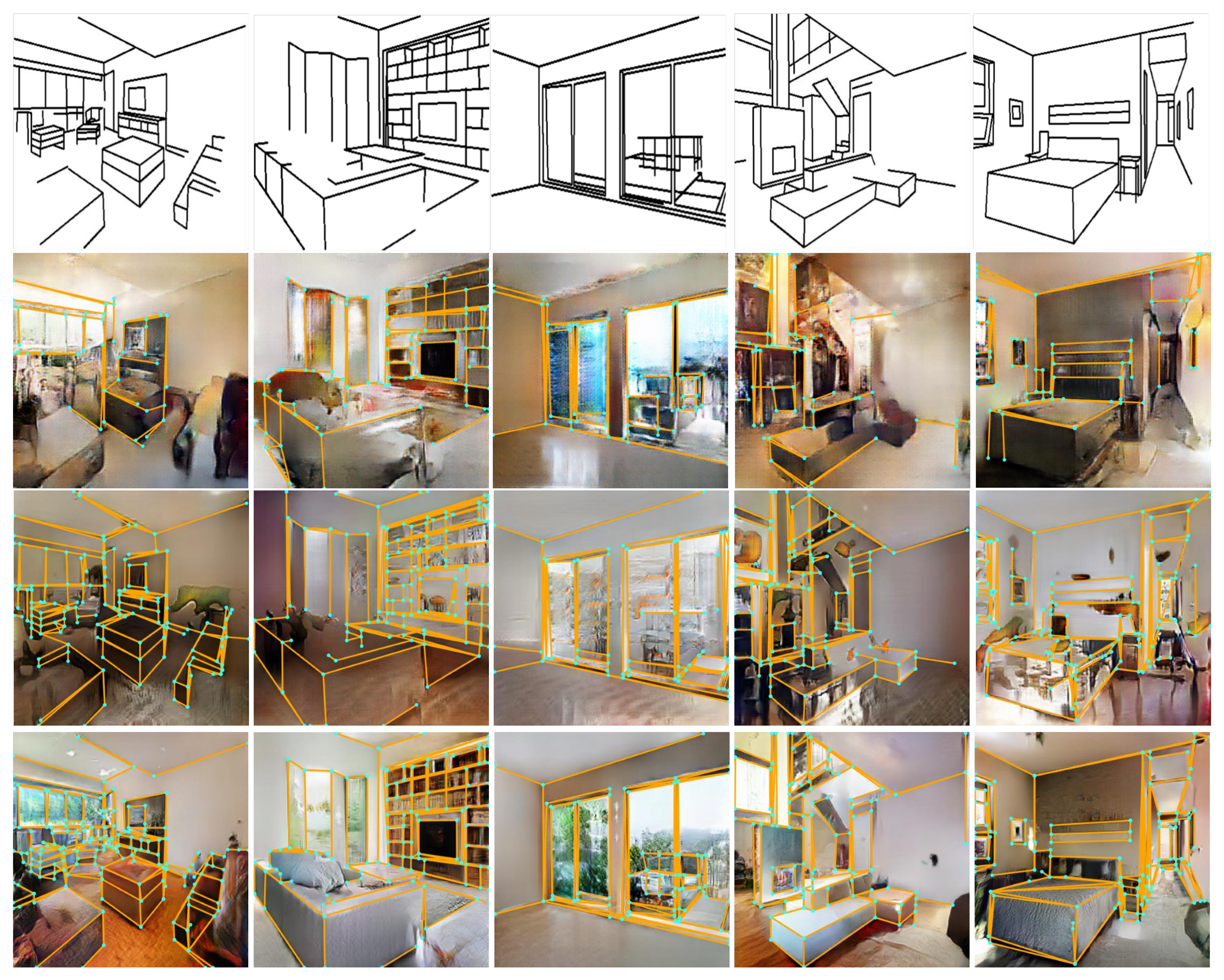}
\caption{Example wireframe detection results on synthesized images. {\bf First row}: Input wireframe. {\bf Second to fourth rows}: Detection results on images generated by pix2pixHD, SPADE, and our method, respectively. Wireframes are detected by the wireframe parser~\cite{zhou2019end}. For fair comparison, no post-processing is done for the parser.
}
\label{fig:detection}
\end{figure}

To this end, we propose a wireframe detection score as a complimentary metric for evaluating the structural integrity in image translation systems. Specifically, we apply the state-of-the-art wireframe parser~\cite{zhou2019end} to detect wireframes from the generated images. The wireframe parser outputs a vectorized wireframe that contains semantically
meaningful and geometrically salient junctions and lines (Fig.~\ref{fig:detection}). To evaluate the results, we follow~\cite{zhou2019end} and use the \emph{structural average precision (sAP)}, which is defined as the area under the precision-recall curve computed from a scored list of detected line segments on all test images. Here, a detected line is considered as a true positive if the distance between the predicted and ground truth end points is within a threshold $\theta$. 

Table~\ref{Tb:translation}(right) reports the sAP scores at $\theta=\{5, 10, 15\}$. 
As one can see, our full model outperforms all other methods. While SPADE also gets relatively high sAP scores by encoding wireframes in all normalization layers, their generated images contain more artifacts.
In the last row of Table~\ref{Tb:translation}(right), we also report sAP scores obtained by applying the same wireframe parser~\cite{zhou2019end} to the corresponding real images. Rather surprisingly, the images generated by our method even achieve higher sAP scores than the real images. After a close inspection of the results, we find that it is mainly because, when labeling wireframe, human annotators tend to miss some salient lines and junctions in the real images. In other words, there are often more salient lines and junctions in real images than those labelled in the ground truth. As a result, the detected wireframes from real images contain more false positives. In the meantime, the input provided to our model is just the annotated wireframes. Our model is able to faithfully preserve such information in the generated images.

\medskip
\noindent{\bf Human studies.} We also conduct a human perception evaluation to compare the quality of generated images between our
method and pix2pixHD, since SPADE results contain more artifacts. We show the ground truth wireframes paired with images generated by our method and pix2pixHD to three workers. The workers are asked to evaluate synthetic images based on fidelity and the alignment to wireframes. They are given unlimited time to choose between our method, pix2pixHD, or ``none'' if both methods fail to generate realistic enough images or preserve the wireframe structure. We use all 422 test images for this evaluation. On average, the preference rates of pix2pixHD, our method, and ``none'' are $3.7\%, 65.1\%, 31.2\%$, respectively. The human study result further proves that our method can not only generate realistic renderings, but also respect the structure information encoded in wireframes.

\subsection{Wireframe Manipulation}

\begin{figure}
\centering
\begin{tabular}{cccc}
\includegraphics[height=0.56in]{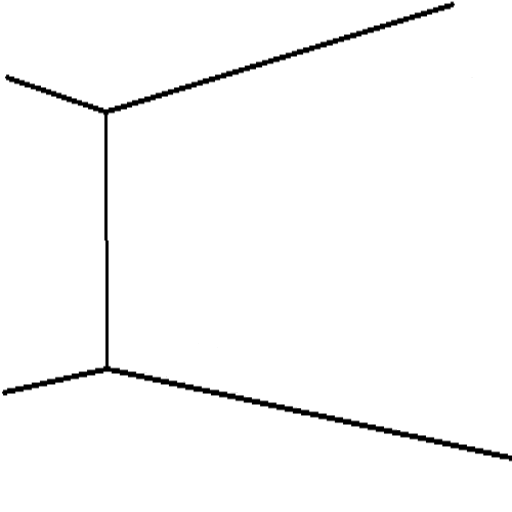}
\includegraphics[height=0.56in]{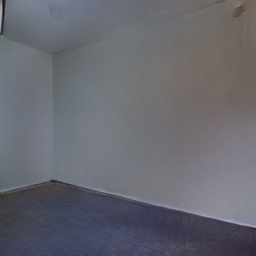}&
\includegraphics[height=0.56in]{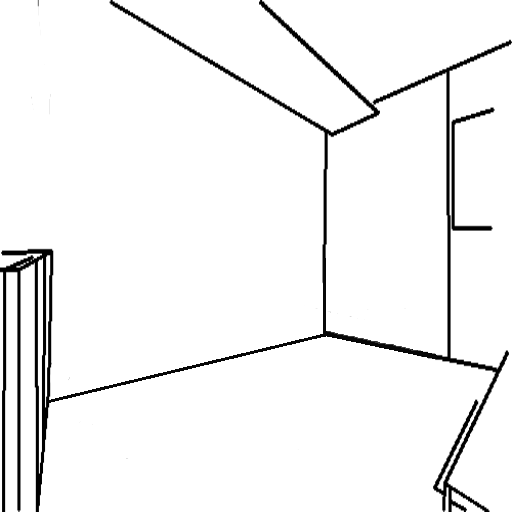}
\includegraphics[height=0.56in]{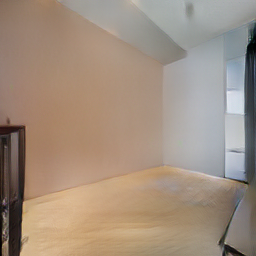}&
\includegraphics[height=0.56in]{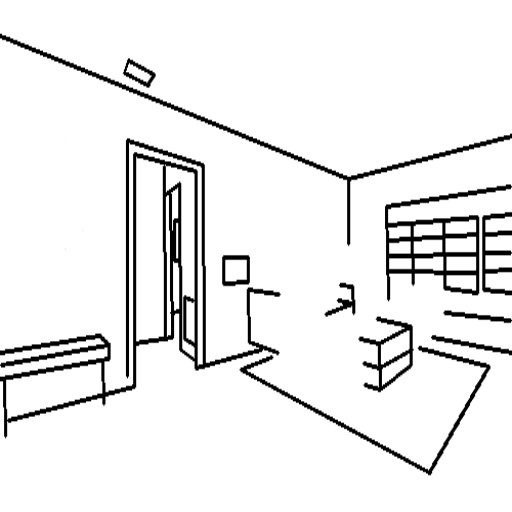}
\includegraphics[height=0.56in]{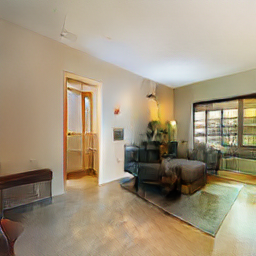}&
\includegraphics[height=0.56in]{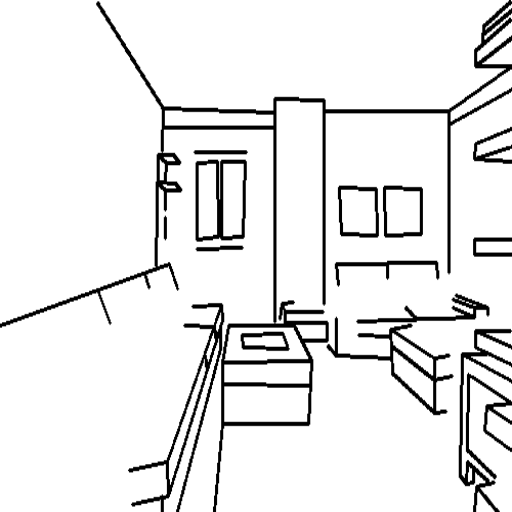}
\includegraphics[height=0.56in]{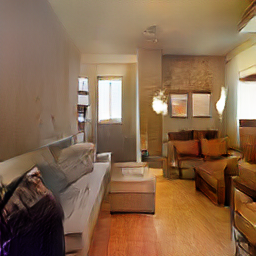}\\
\includegraphics[height=0.56in]{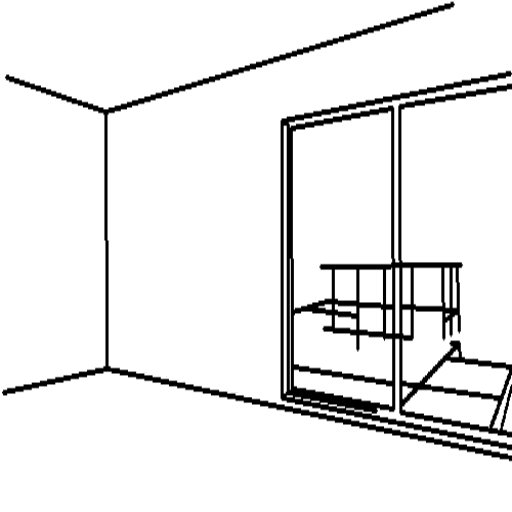}
\includegraphics[height=0.56in]{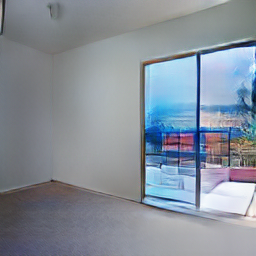}&
\includegraphics[height=0.56in]{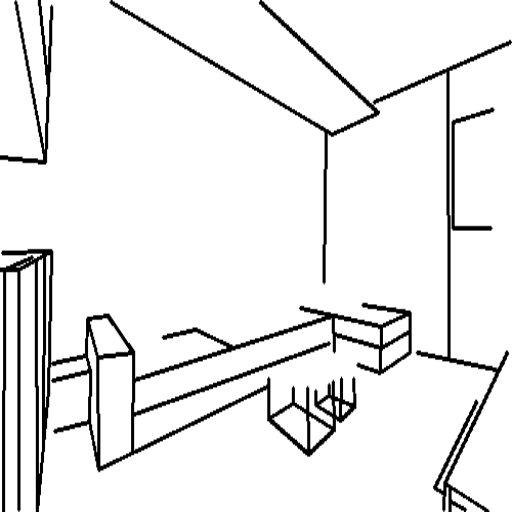}
\includegraphics[height=0.56in]{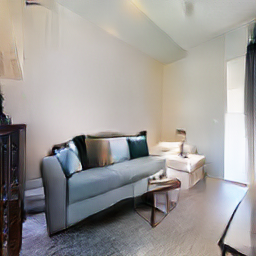}&
\includegraphics[height=0.56in]{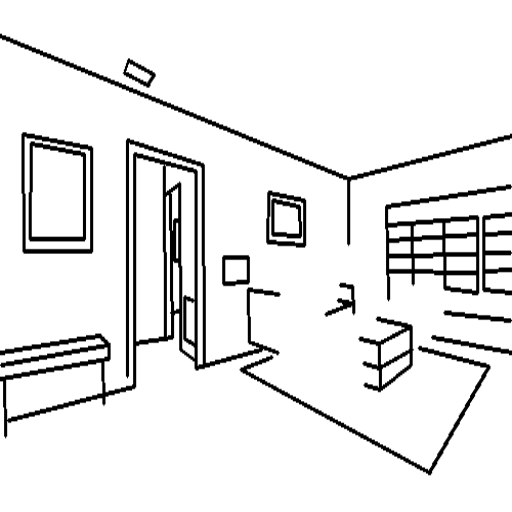}
\includegraphics[height=0.56in]{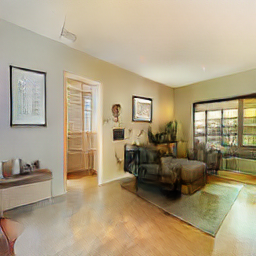}&
\includegraphics[height=0.56in]{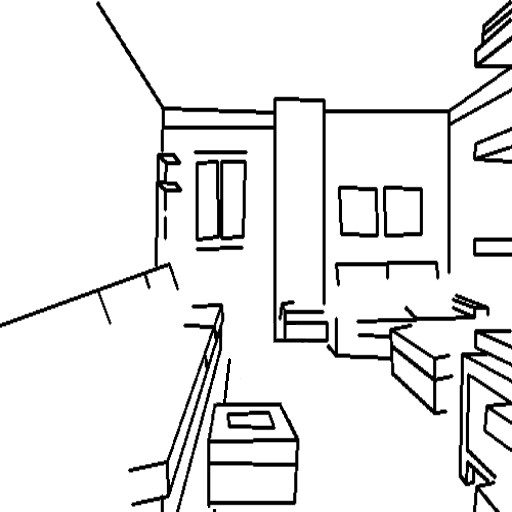}
\includegraphics[height=0.56in]{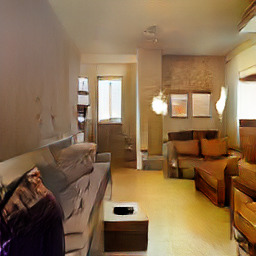}
\\
(a) & (b) & (c) & (d)
\end{tabular}
\caption{Example wireframe manipulation results. {\bf Odd columns}: Input wireframes; {\bf Even columns}: Images generated by our model. Modifications include {\bf (a)} adding exterior doors, {\bf (b)} adding furniture set, {\bf (c)} adding wall decoration, and {\bf (d)} relocating a table in front of the sofa.}
\label{fig:manipulation}
\end{figure}

To provide additional insight to our model, and also to illustrate the potential use of our method in a realistic design application setting, we incrementally modify the input wireframes and check whether the generated scene images are updated in a meaningful way. As shown in Fig.~\ref{fig:manipulation}, we manually edit some lines/junctions in the original wireframe. 
The results show that our model captures the changes in the wireframe and updates the generated image in a consistent fashion.

\subsection{Color Guided Rendering}
The previous experiments focus on using wireframes as the only input to generate color images. In this section, we introduce a color guided image generation process, which allows users to provide additional input that specifies the color theme of the rendered scene. A color guided model enables finer control of the rendering process and can be more applicable in real world design settings.
Specifically, we encode a RGB color histogram (a $256\times 3$ dimensional vector) as additional input into our joint representation. The input color histogram is first normalized to the range $[0,1]$ then converted into the same size as the input via a linear layer. The input to our model is the concatenation of the wireframe and the transformed color histogram. On the decoder side, a linear layer and a sigmoid layer are used to reconstruct the color histogram from the synthesized image with a $\ell_1$ loss. 
With our model, users can apply any color histogram from existing designs to an input wireframe to get a rendering result with desired color or style. 

\begin{figure}[t]
\centering
\includegraphics[width=0.90\linewidth]{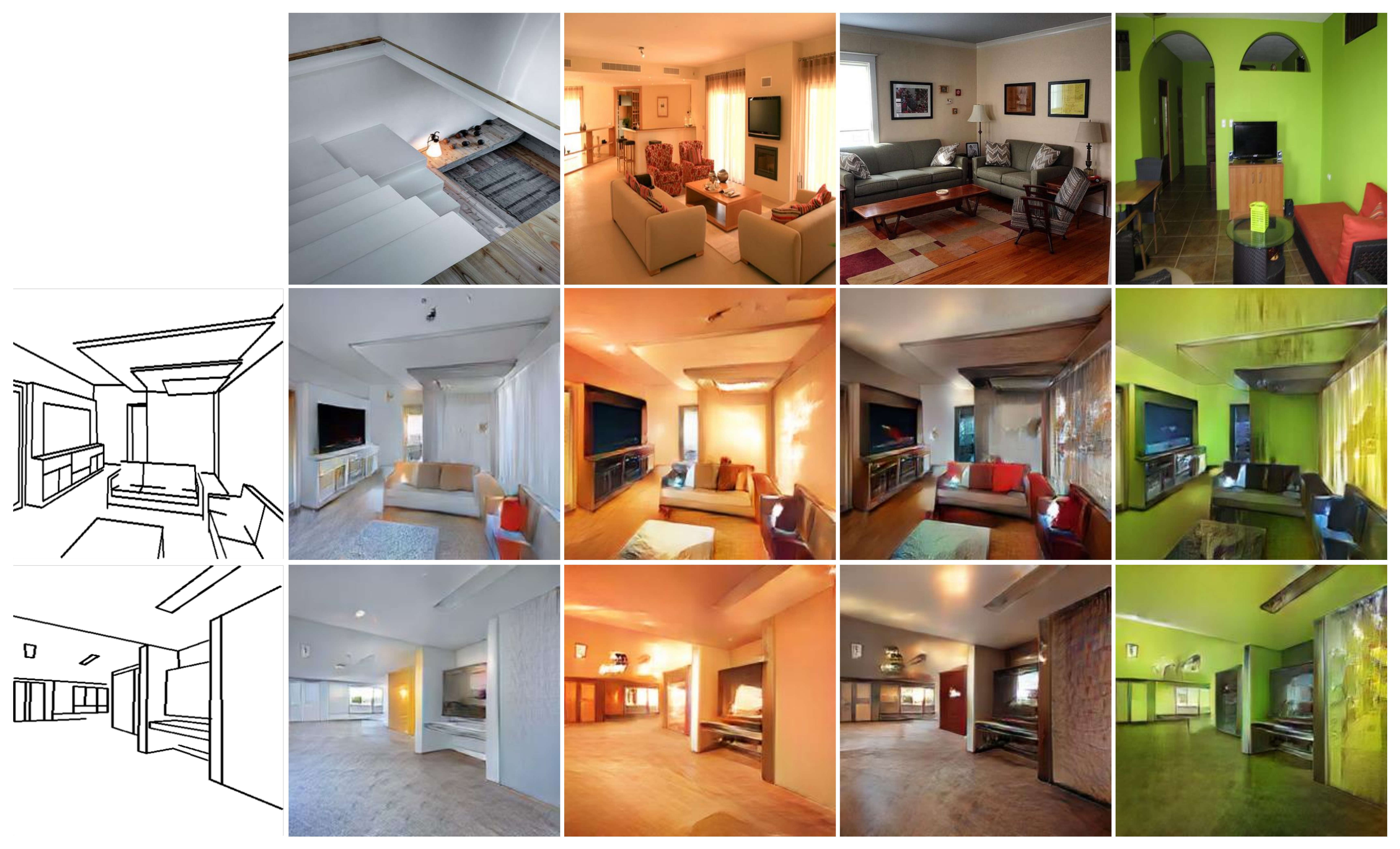}
\caption{Example color guided rendering results. First column shows the input wireframes; first row shows the real images providing color guidance; the rest are images generated by our model.}
\label{fig:color_guide}
\end{figure}

Fig.~\ref{fig:color_guide} shows the results of the color guided generation process. Note how our model adapts to very different color schemes while maintaining consistent room layout and configuration:
Even with unusual rust and emerald colors, our model responds well, for example, by replacing regular walls with crystal ones. This demonstrates the potential of our method in interactive and customized design.

\subsection{Discussion}
\begin{figure}[t]
\centering
\includegraphics[width=0.90\linewidth]{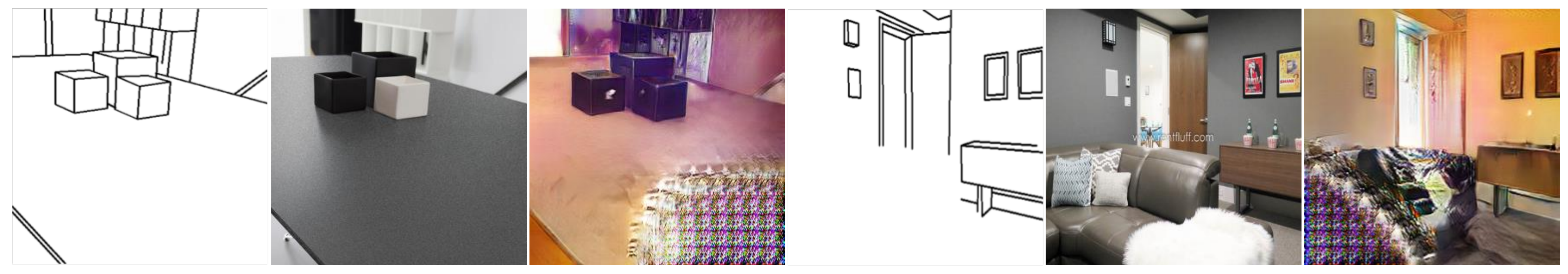}
\caption{Failure examples generated by our model.}
\label{fig:fail}
\end{figure}

\noindent{\bf Failure cases.} While our model is able to capture structure information well, the input wireframes can be sparse and contain little semantic information such as objects. As shown in Fig.~\ref{fig:fail}, when there is little wireframe information provided to the model, especially in the corner part, our model sometimes fails to generate visually meaningful results. We expect to mitigate this issue by training on a larger dataset containing more diverse images and wireframes, and providing other semantic inputs such as types of furniture to make the learning task easier.

\smallskip
\noindent{\bf Extensions.}
Our joint representation learning framework is general and may also benefit other image synthesis tasks. In fact, we have conducted preliminary experiments on the noise-to-image generation task, in which our model is trained to simultaneously generate paired scene image and wireframe from a noise input using the joint representation. We have obtained improved results compared to a baseline which generates the scene image only. More details are provided in the supplementary material.

\section{Conclusion}
In this paper, we study a new image synthesis task for design applications in which the input is a wireframe representation of a scene. By learning the joint representation in the shared latent space of wireframes and images, our wireframe render generates photo-realistic scene images with high structural integrity. In the future, we plan to extend our model to a wider range of design scenarios and consider semantic constraints alongside with structural constraints. To exploit the flexibility of our model, we will also investigate generating image renderings directly from computer-aided design (CAD) models (in vector-graphics format) with different viewpoint projections.

\clearpage
%
%
\bibliographystyle{splncs04}
\bibliography{egbib}

\begin{thebibliography}{10}
\providecommand{\url}[1]{\texttt{#1}}
\providecommand{\urlprefix}{URL }
\providecommand{\doi}[1]{https://doi.org/#1}

\bibitem{chen2017photographic}
Chen, Q., Koltun, V.: Photographic image synthesis with cascaded refinement
  networks. In: ICCV. pp. 1511--1520 (2017)

\bibitem{chen2018sketchygan}
Chen, W., Hays, J.: Sketchygan: Towards diverse and realistic sketch to image
  synthesis. In: CVPR. pp. 9416--9425 (2018)

\bibitem{chen2016infogan}
Chen, X., Duan, Y., Houthooft, R., Schulman, J., Sutskever, I., Abbeel, P.:
  Infogan: Interpretable representation learning by information maximizing
  generative adversarial nets. In: NIPS. pp. 2172--2180 (2016)

\bibitem{ChenLCG19}
Chen, Y., Li, W., Chen, X., Gool, L.V.: Learning semantic segmentation from
  synthetic data: {A} geometrically guided input-output adaptation approach.
  In: CVPR. pp. 1841--1850 (2019)

\bibitem{choi2018stargan}
Choi, Y., Choi, M., Kim, M., Ha, J.W., Kim, S., Choo, J.: Stargan: Unified
  generative adversarial networks for multi-domain image-to-image translation.
  In: CVPR. pp. 8789--8797 (2018)

\bibitem{dosovitskiy2016generating}
Dosovitskiy, A., Brox, T.: Generating images with perceptual similarity metrics
  based on deep networks. In: NIPS. pp. 658--666 (2016)

\bibitem{dumoulin2016adversarially}
Dumoulin, V., Belghazi, I., Poole, B., Lamb, A., Arjovsky, M., Mastropietro,
  O., Courville, A.C.: Adversarially learned inference. In: ICLR (2017)

\bibitem{goodfellow2014generative}
Goodfellow, I., Pouget-Abadie, J., Mirza, M., Xu, B., Warde-Farley, D., Ozair,
  S., Courville, A., Bengio, Y.: Generative adversarial nets. In: NIPS. pp.
  2672--2680 (2014)

\bibitem{gulrajani2017improved}
Gulrajani, I., Ahmed, F., Arjovsky, M., Dumoulin, V., Courville, A.C.: Improved
  training of wasserstein gans. In: NIPS. pp. 5767--5777 (2017)

\bibitem{he2016deep}
He, K., Zhang, X., Ren, S., Sun, J.: Deep residual learning for image
  recognition. In: CVPR. pp. 770--778 (2016)

\bibitem{heusel2017gans}
Heusel, M., Ramsauer, H., Unterthiner, T., Nessler, B., Hochreiter, S.: Gans
  trained by a two time-scale update rule converge to a local nash equilibrium.
  In: NIPS. pp. 6626--6637 (2017)

\bibitem{HoffmanTPZISED18}
Hoffman, J., Tzeng, E., Park, T., Zhu, J., Isola, P., Saenko, K., Efros, A.A.,
  Darrell, T.: Cycada: Cycle-consistent adversarial domain adaptation. In:
  ICML. pp. 1994--2003 (2018)

\bibitem{huang2018learning}
Huang, K., Wang, Y., Zhou, Z., Ding, T., Gao, S., Ma, Y.: Learning to parse
  wireframes in images of man-made environments. In: CVPR. pp. 626--635 (2018)

\bibitem{huang2018multimodal}
Huang, X., Liu, M.Y., Belongie, S., Kautz, J.: Multimodal unsupervised
  image-to-image translation. In: ECCV. pp. 172--189 (2018)

\bibitem{ioffe2015batch}
Ioffe, S., Szegedy, C.: Batch normalization: Accelerating deep network training
  by reducing internal covariate shift. In: ICML. pp. 448--456 (2015)

\bibitem{isola2017image}
Isola, P., Zhu, J.Y., Zhou, T., Efros, A.A.: Image-to-image translation with
  conditional adversarial networks. In: CVPR. pp. 1125--1134 (2017)

\bibitem{johnson2016perceptual}
Johnson, J., Alahi, A., Fei-Fei, L.: Perceptual losses for real-time style
  transfer and super-resolution. In: ECCV. pp. 694--711. Springer (2016)

\bibitem{karras2017progressive}
Karras, T., Aila, T., Laine, S., Lehtinen, J.: Progressive growing of gans for
  improved quality, stability, and variation. In: ICLR (2018)

\bibitem{karras2018style}
Karras, T., Laine, S., Aila, T.: A style-based generator architecture for
  generative adversarial networks. In: CVPR. pp. 4401--4410 (2019)

\bibitem{kim2017learning}
Kim, T., Cha, M., Kim, H., Lee, J.K., Kim, J.: Learning to discover
  cross-domain relations with generative adversarial networks. In: ICML. pp.
  1857--1865. JMLR. org (2017)

\bibitem{kingma2014adam}
Kingma, D.P., Ba, J.: Adam: {A} method for stochastic optimization. In: ICLR
  (2015)

\bibitem{ledig2017photo}
Ledig, C., Theis, L., Husz{\'a}r, F., Caballero, J., Cunningham, A., Acosta,
  A., Aitken, A., Tejani, A., Totz, J., Wang, Z., et~al.: Photo-realistic
  single image super-resolution using a generative adversarial network. In:
  CVPR. pp. 4681--4690 (2017)

\bibitem{LeeTHSY18}
Lee, H., Tseng, H., Huang, J., Singh, M., Yang, M.: Diverse image-to-image
  translation via disentangled representations. In: ECCV. pp. 36--52 (2018)

\bibitem{li2017alice}
Li, C., Liu, H., Chen, C., Pu, Y., Chen, L., Henao, R., Carin, L.: Alice:
  Towards understanding adversarial learning for joint distribution matching.
  In: NIPS. pp. 5495--5503 (2017)

\bibitem{liu2017unsupervised}
Liu, M.Y., Breuel, T., Kautz, J.: Unsupervised image-to-image translation
  networks. In: NIPS. pp. 700--708 (2017)

\bibitem{liu2019learning}
Liu, X., Yin, G., Shao, J., Wang, X., Li, H.: Learning to predict
  layout-to-image conditional convolutions for semantic image synthesis. arXiv
  preprint arXiv:1910.06809  (2019)

\bibitem{long2015fully}
Long, J., Shelhamer, E., Darrell, T.: Fully convolutional networks for semantic
  segmentation. In: CVPR. pp. 3431--3440 (2015)

\bibitem{LuWTT18}
Lu, Y., Wu, S., Tai, Y., Tang, C.: Image generation from sketch constraint
  using contextual {GAN}. In: ECCV. pp. 213--228 (2018)

\bibitem{ma2018gan}
Ma, S., Fu, J., Wen~Chen, C., Mei, T.: Da-gan: Instance-level image translation
  by deep attention generative adversarial networks. In: CVPR. pp. 5657--5666
  (2018)

\bibitem{maas2013rectifier}
Maas, A.L., Hannun, A.Y., Ng, A.Y.: Rectifier nonlinearities improve neural
  network acoustic models. In: ICML (2013)

\bibitem{mao2017least}
Mao, X., Li, Q., Xie, H., Lau, R.Y., Wang, Z., Paul~Smolley, S.: Least squares
  generative adversarial networks. In: ICCV. pp. 2794--2802 (2017)

\bibitem{mirza2014conditional}
Mirza, M., Osindero, S.: Conditional generative adversarial nets. arXiv
  preprint arXiv:1411.1784  (2014)

\bibitem{MoCS19}
Mo, S., Cho, M., Shin, J.: Instagan: Instance-aware image-to-image translation.
  In: ICLR (2019)

\bibitem{MurezKKRK18}
Murez, Z., Kolouri, S., Kriegman, D.J., Ramamoorthi, R., Kim, K.: Image to
  image translation for domain adaptation. In: CVPR. pp. 4500--4509 (2018)

\bibitem{odena2016deconvolution}
Odena, A., Dumoulin, V., Olah, C.: Deconvolution and checkerboard artifacts.
  Distill  \textbf{1}(10), ~e3 (2016)

\bibitem{odena2017conditional}
Odena, A., Olah, C., Shlens, J.: Conditional image synthesis with auxiliary
  classifier gans. In: ICML. pp. 2642--2651 (2017)

\bibitem{park2019semantic}
Park, T., Liu, M.Y., Wang, T.C., Zhu, J.Y.: Semantic image synthesis with
  spatially-adaptive normalization. In: CVPR. pp. 2337--2346 (2019)

\bibitem{qi2018semi}
Qi, X., Chen, Q., Jia, J., Koltun, V.: Semi-parametric image synthesis. In:
  CVPR. pp. 8808--8816 (2018)

\bibitem{ronneberger2015u}
Ronneberger, O., Fischer, P., Brox, T.: U-net: Convolutional networks for
  biomedical image segmentation. In: International Conference on Medical image
  computing and computer-assisted intervention. pp. 234--241. Springer (2015)

\bibitem{russakovsky2015imagenet}
Russakovsky, O., Deng, J., Su, H., Krause, J., Satheesh, S., Ma, S., Huang, Z.,
  Karpathy, A., Khosla, A., Bernstein, M., et~al.: Imagenet large scale visual
  recognition challenge. International Journal of Computer Vision
  \textbf{115}(3),  211--252 (2015)

\bibitem{SangkloyLFYH17}
Sangkloy, P., Lu, J., Fang, C., Yu, F., Hays, J.: Scribbler: Controlling deep
  image synthesis with sketch and color. In: CVPR. pp. 6836--6845 (2017)

\bibitem{shi2016real}
Shi, W., Caballero, J., Husz{\'a}r, F., Totz, J., Aitken, A.P., Bishop, R.,
  Rueckert, D., Wang, Z.: Real-time single image and video super-resolution
  using an efficient sub-pixel convolutional neural network. In: CVPR. pp.
  1874--1883 (2016)

\bibitem{simonyan2014very}
Simonyan, K., Zisserman, A.: Very deep convolutional networks for large-scale
  image recognition. arXiv preprint arXiv:1409.1556  (2014)

\bibitem{szegedy2016rethinking}
Szegedy, C., Vanhoucke, V., Ioffe, S., Shlens, J., Wojna, Z.: Rethinking the
  inception architecture for computer vision. In: CVPR. pp. 2818--2826 (2016)

\bibitem{wang2018perceptual}
Wang, C., Xu, C., Wang, C., Tao, D.: Perceptual adversarial networks for
  image-to-image transformation. IEEE Transactions on Image Processing
  \textbf{27}(8),  4066--4079 (2018)

\bibitem{wang2018high}
Wang, T.C., Liu, M.Y., Zhu, J.Y., Tao, A., Kautz, J., Catanzaro, B.:
  High-resolution image synthesis and semantic manipulation with conditional
  gans. In: CVPR. pp. 8798--8807 (2018)

\bibitem{wang2016generative}
Wang, X., Gupta, A.: Generative image modeling using style and structure
  adversarial networks. In: ECCV. pp. 318--335. Springer (2016)

\bibitem{wang2004image}
Wang, Z., Bovik, A.C., Sheikh, H.R., Simoncelli, E.P., et~al.: Image quality
  assessment: from error visibility to structural similarity. IEEE Transactions
  on Image Processing  \textbf{13}(4),  600--612 (2004)

\bibitem{wang2003multiscale}
Wang, Z., Simoncelli, E.P., Bovik, A.C.: Multiscale structural similarity for
  image quality assessment. In: The 37th Asilomar Conference on Signals,
  Systems \& Computers. vol.~2, pp. 1398--1402 (2003)

\bibitem{xue2019learning}
Xue, N., Bai, S., Wang, F., Xia, G.S., Wu, T., Zhang, L.: Learning attraction
  field representation for robust line segment detection. In: CVPR. pp.
  1595--1603 (2019)

\bibitem{yi2017dualgan}
Yi, Z., Zhang, H., Tan, P., Gong, M.: Dualgan: Unsupervised dual learning for
  image-to-image translation. In: ICCV. pp. 2849--2857 (2017)

\bibitem{zhang2017stackgan++}
Zhang, H., Xu, T., Li, H., Zhang, S., Wang, X., Huang, X., Metaxas, D.:
  Stackgan++: Realistic image synthesis with stacked generative adversarial
  networks. arXiv preprint arXiv:1710.10916  (2017)

\bibitem{zhang2017image}
Zhang, H., Sindagi, V., Patel, V.M.: Image de-raining using a conditional
  generative adversarial network. arXiv preprint arXiv:1701.05957  (2017)

\bibitem{zhang2018unreasonable}
Zhang, R., Isola, P., Efros, A.A., Shechtman, E., Wang, O.: The unreasonable
  effectiveness of deep features as a perceptual metric. In: CVPR. pp. 586--595
  (2018)

\bibitem{zhou2019end}
Zhou, Y., Qi, H., Ma, Y.: End-to-end wireframe parsing. In: ICCV 2019 (2019)

\bibitem{Zhou19}
Zhou, Y., Qi, H., Zhai, S., Sun, Q., Chen, Z., Wei, L.Y., Ma, Y.: Learning to
  reconstruct 3d manhattan wireframes from a single image. In: ICCV (2019)

\bibitem{zhu2017unpaired}
Zhu, J.Y., Park, T., Isola, P., Efros, A.A.: Unpaired image-to-image
  translation using cycle-consistent adversarial networks. In: ICCV. pp.
  2223--2232 (2017)

\end{thebibliography}

\newpage
\appendix

\section{Joint Wireframe and Image Synthesis}

Our joint representation learning framework is general and may also benefit other image synthesis tasks. In this section, we report preliminary results on extending this framework to the noise-to-image task.  Recently, coarse-to-fine multi-scale models~\cite{chen2017photographic,karras2017progressive,wang2018high,zhang2017stackgan++} have been shown to produce visually pleasing results. However, such models often rely on having a large training set and do not explicitly take structural integrity into consideration. 

\begin{figure}[!ht]
\centering
  \includegraphics[width=0.85\linewidth]{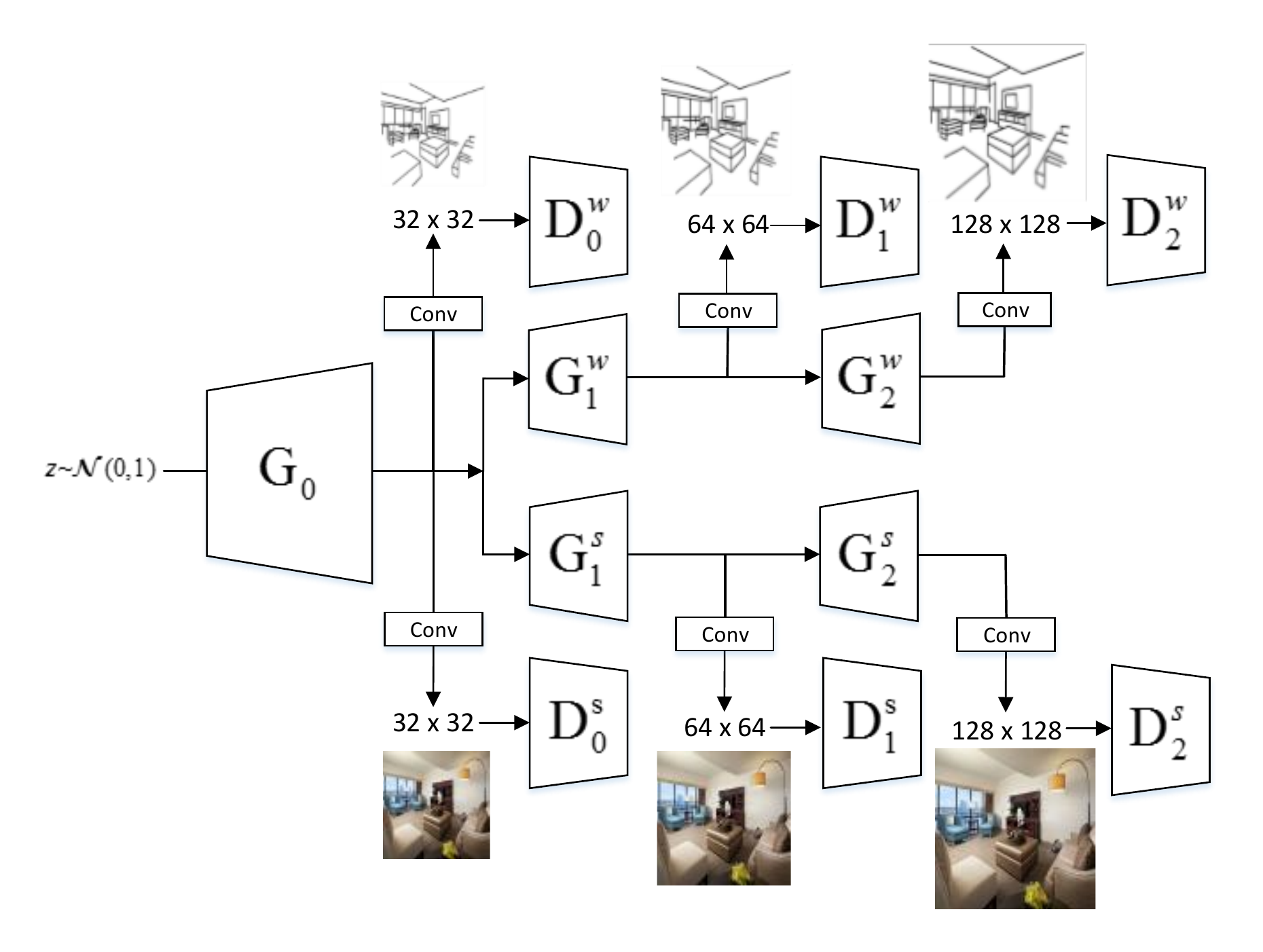}
  \caption{Network architecture of our GAN model for joint wireframe and image generation. The backbone generators and discriminators are the same as StackGANv2~\cite{zhang2017stackgan++}. 
  }
\label{fig:framework_generation}
\vspace{-3mm}
\end{figure}

In this work, we choose the state-of-art StackGANv2~\cite{zhang2017stackgan++} model as the baseline model for our experiments. In the baseline model, the generator takes a random noise vector as input and output an image. 
Instead of generating images only, we propose a GAN model to generate images and their corresponding wireframes simultaneously. 
An illustration of our proposed GAN model with joint representation learning can be found in Fig.~\ref{fig:framework_generation}.
Unlike the original StackGANv2 model, we first map the input noise vector to a shared latent space of wireframe and image through the first generator $G_0$, then two separate branches of coarse-to-fine generators take the joint representation and generate wireframes and images, respectively. Although we do not have explicit supervision upon the joint representation, the wireframe-based adversarial learning guarantees that the learned representation contains enough structure information. 

Note that our GAN model for image and wireframe generation does not require paired wireframes and images during training, as it uses separate discriminators for wireframes and images.
Thus, we can potentially use wireframes and images from different sources, which makes the model scalable to much larger datasets.

Following~\cite{zhang2017stackgan++}, our GAN objective consists of two parts: the traditional adversarial loss and a color-consistency regularization term. Since we are generating both wireframe and image, we also apply a structure-consistency regularization term to the generated wireframes. More specifically, the adversarial objective for the $i$th generator $G_i$ and the $i$th discriminator $D_i$ is defined as
\begin{equation}
\begin{split}
\max_{\theta_D}\min_{\theta_G}  \mathcal{L}^{\text{adv}}_i
= & \mathbb{E}_{x_i}[\log D_i^w(x_i)] + \mathbb{E}_{z_i^w}\log (1 - D_i^w(G_i^w(z_i^w)))]\\
&+ \mathbb{E}_{y_i}[\log D_i^s(y_i)] + \mathbb{E}_{z_i^s}\log (1 - D_i^s(G_i^s(z_i^s)))],
\end{split}
\end{equation}
where $z_i$ is the input to the $i$th generator, $x$ and $y$ represent real wireframes and images, respectively. The superscript indicates the index of the generator/discriminator branch and $G_0$ is shared by both branches. 

Given a mini-batch of $N$ generated wireframes $\hat{x}^n_i$ and images $\hat{y}^n_i$ at the $i$th scale, the color- and structure-consistency regularization term is defined as
\begin{equation}
\begin{split}
\mathcal{L}^{\text{con}}_i = \frac{1}{N}\sum_{n=1}^N &||\lambda_1 \bm{\mu}_{\hat{x}^n_i} - \bm{\mu}_{\hat{x}^n_{i-1}}||_2^2 + ||\lambda_2 \bm{\Sigma}_{\hat{x}^n_i} - \bm{\Sigma}_{\hat{x}^n_{i-1}}||_2^2\\
&+ ||\lambda_1 \bm{\mu}_{\hat{y}^n_i} - \bm{\mu}_{\hat{y}^n_{i-1}}||_2^2 + ||\lambda_2 \bm{\Sigma}_{\hat{y}^n_i} - \bm{\Sigma}_{\hat{y}^n_{i-1}}||_2^2,
\end{split}
\end{equation}
where $\bm{\mu}$ and $\bm{\Sigma}$ represent the mean and covariance of pixel values of the given wireframe or image. 

During the training of each discriminator in our model, only the adversarial loss $\mathcal{L}^{\text{adv}}$ is applied. When we train the $i$th generator, the total loss is the sum of the adversarial loss and the consistency regularization loss, \textit{i.e.}, $\mathcal{L}^{G}_i = \mathcal{L}^{\text{adv}}_i + \alpha_{\text{con}} \mathcal{L}^{\text{con}}_i$, where $\alpha_{\text{con}}$ is the scaling factor that controls the relative influence of the two loss terms.

\subsection{Implementation Details}

Our proposed GAN model is built upon a StackGANv2~\cite{zhang2017stackgan++} backbone. After the shared generator $G_0$ which maps the input vector $z$ to a joint embedding, the wireframe generator and image generator use separate coarse-to-fine generators $G_1^w, G_1^s$ and $G_2^w, G_2^s$ to generate wireframes and images at different scales. Before generating each wireframe/image, the learned features will go through a $3 \times 3$ convolution block including batch normalization and relu activation, then followed by a $7\times7$ convolution and tanh activation to generate the wireframe or image. We set $\lambda_1 = 1$, $\lambda_2 = 5$ and $\alpha_{\text{con}}=50$ to be consistent with the original StackGANv2. The training is done by Adam optimizer with fixed learning rate $2e-3$. The batchsize is $64$ and the maximum number of training epochs is $500$. No LSGAN loss is applied during training.

The data pre-processing is the same as in the image translation experiments except that we do not apply color jitter augmentation for GAN training. During inference, only the highest resolution images (we use $128 \times 128$ in joint synthesis experiments) are evaluated.

\subsection{Experiment Results}

\begin{figure}[!t]
\centering
\includegraphics[width=0.99\linewidth]{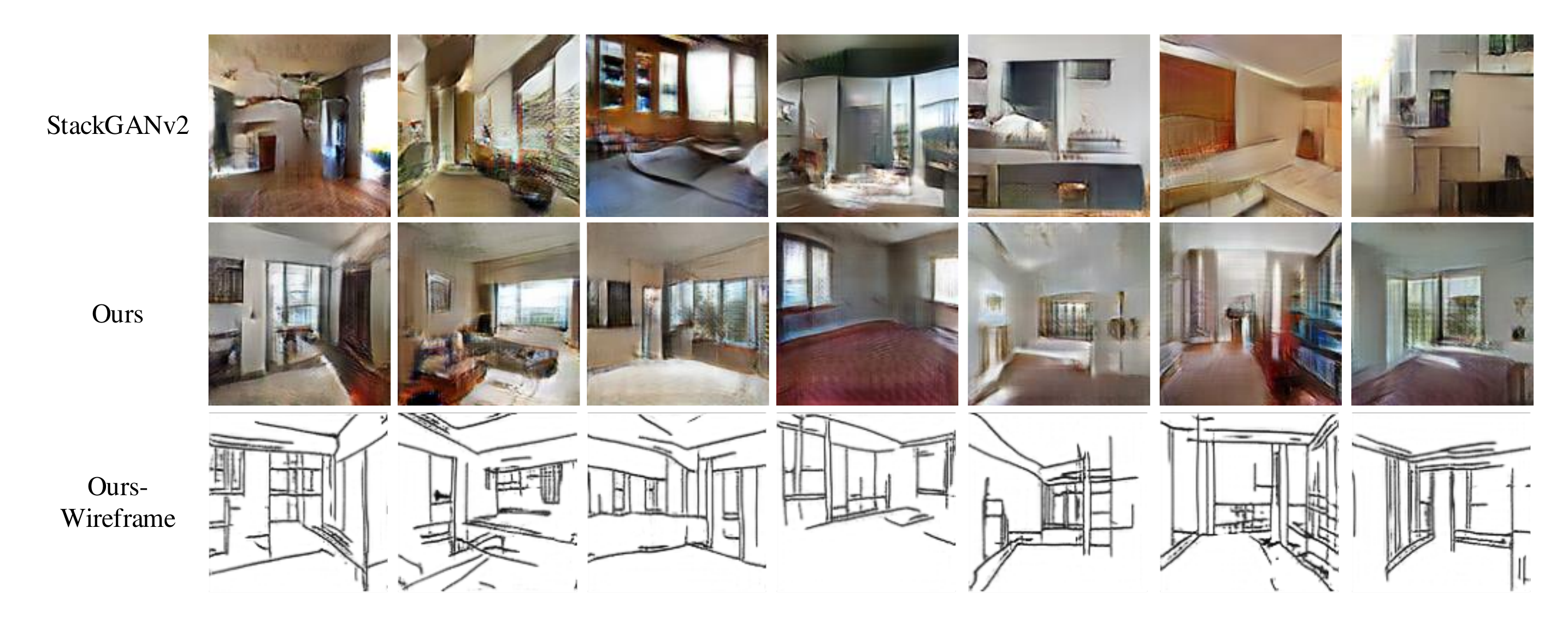}
\caption{Qualitative comparisons of image generation models. The first row contains generated images by the baseline StackGANv2~\cite{zhang2017stackgan++} model. The second and third rows are paired images and wireframes generated by our model.}
\label{fig:result_generation}
\end{figure}

Fig.~\ref{fig:result_generation} shows example image synthesis results of StackGANv2 and our model. 
As one can see, our model generates images with room layouts which are more geometrically meaningful and better align with the typical layout of real rooms. It is also worth noting that the wireframes generated by our model align quite well with the images, despite that fact that no direct supervision is provided w.r.t. the alignment. This is a strong indication of the effectiveness of the joint representation learning module.

\begin{table}[t]
\centering
\caption{Quantitative comparisons between image generation models. The inception score of the real images in the test set is provided as reference.}
\label{Tb:generation}
\begin{tabular}{l|c|c}
\hline
Method & IS$\uparrow$ & FID$\downarrow$  \\
\hline\hline
StackGANv2~\cite{zhang2017stackgan++} & $2.92$ & $\bm{49.76}$\\
\hline
Ours & $\bm{3.08}$ & ${50.96}$  \\
\hline
GT & ${3.21}$ & -   \\
\hline
\end{tabular}
\end{table}

Table~\ref{Tb:generation} reports quantitative comparison results between the baseline StackGANv2 model and our GAN model. Specifically, we randomly generate 500 images for each model, then calculate the IS and FID scores based on the generated images and real images in the test set. 
Here we note that, the focus of our work is more on the geometric constraints and structural integrity of the generated images. However, existing GAN metrics, such as the IS and FID scores, are mainly designed to measure the perceptual quality and the diversity of the generated images, and cannot well capture the structure information. 
While preserving structural integrity in image synthesis remains a challenge for current GAN models, we hope that our proposed model and preliminary results provide some useful insight for future research. We also expect that our model can be improved by training with larger datasets from multiple sources and utilizing advanced GAN models.

\section{Additional Wireframe Rendering Results}

\begin{figure}[h]
\centering
\includegraphics[width=0.99\linewidth]{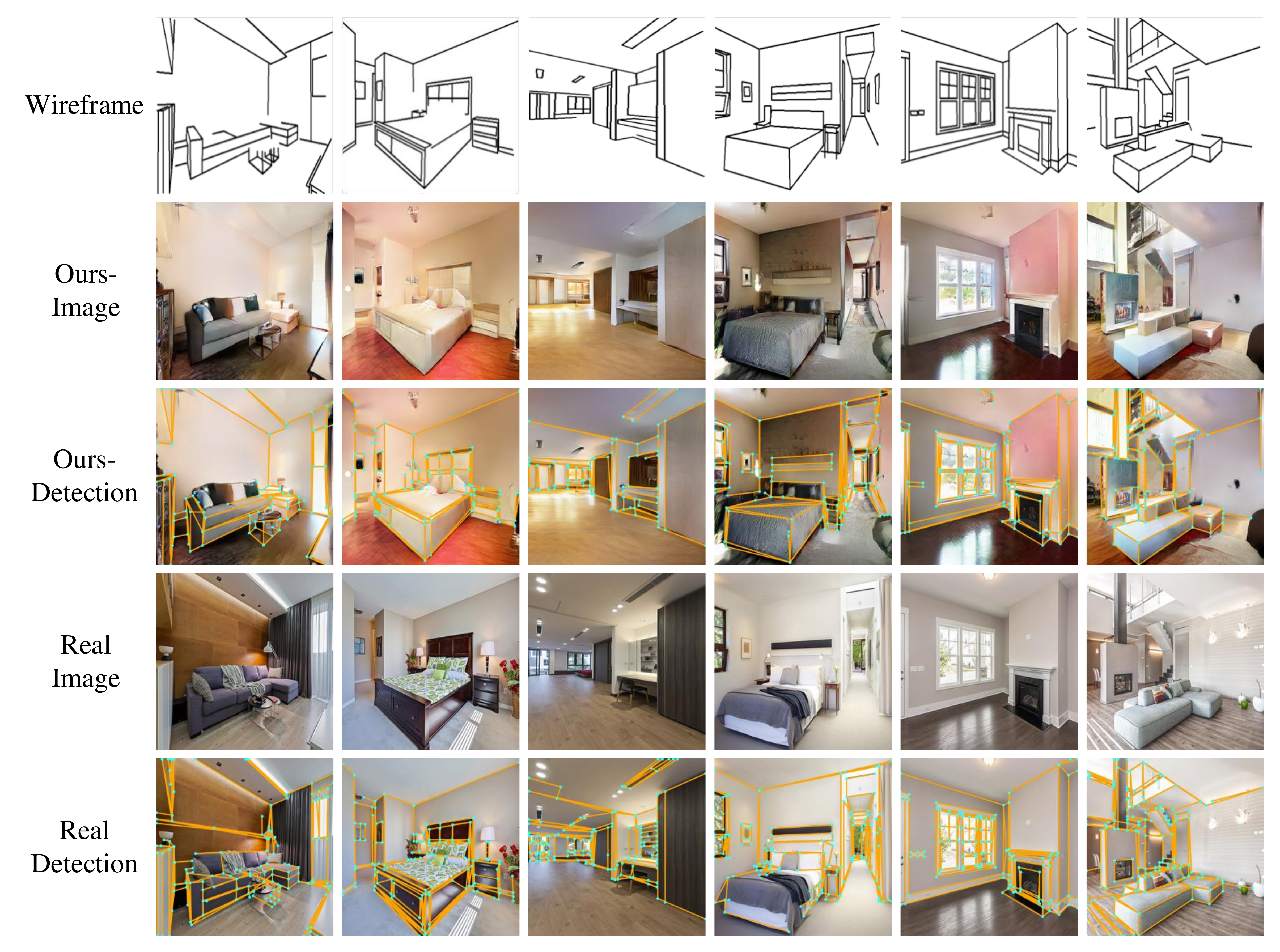}
  \caption{Additional wireframe-to-image translation results. The first row is the input wireframe; second and third rows are images generated by our model and the corresponding wireframe detection results; the rest are real images and the corresponding detection results. We use wireframe parser from~\cite{zhou2019end} to detect wireframes from synthetic/real images. For fair comparison, no post-processing is done for the wireframe parser.}
\label{fig:result_supp}
\vspace{-5mm}
\end{figure}

We provide more wireframe-to-image translation results and wireframe detection results in Fig.~\ref{fig:result_supp}. Note the structural similarity between our synthesized images and the real images. Also, by comparing the wireframe detection results from synthetic images with real images, we can observe that wireframes detected from real images contain more false positives. This may explain why our synthetic images achieves higher sAP scores than real images.

\end{document}